\documentclass[10pt,twocolumn]{article}
\usepackage[utf8]{inputenc}
\usepackage[T1]{fontenc}
\usepackage{arxiv}
\usepackage[numbers,sort&compress]{natbib}

\renewcommand{\shorttitle}{Passive AI physical state verification for liquid handlers}
\preprintdate{October 1, 2026}

\title{A Passive AI System for Verifying Physical State on Automated Liquid Handlers}

\author[1,+]{Junqiong Joanne Qiu}
\author[2,+]{Zeckria Kamrany}
\author[2,+]{Ananya Anand}
\author[3,+]{Emma Vidal}
\author[2,+]{Yunjia Johanna Bai}
\author[3,+]{Nathan S Chen}
\author[2,+]{Sean Son}
\author[1]{Varada Abhyankar}
\author[1]{Taryn Jakub}
\author[1,2,*]{Eleazar Eskin}
\affil[1]{Department of Computational Medicine, David Geffen School of Medicine, UCLA, Los Angeles, CA, USA}
\affil[2]{Department of Computer Science, UCLA, Los Angeles, CA, USA}
\affil[3]{Department of Mathematics, UCLA, Los Angeles, CA, USA}
\affil[*]{Corresponding author}
\affil[+]{These authors contributed equally to this work}


\preprintabstract{Automated liquid handlers execute digital protocols, but operators must assemble the deck and confirm that the physical setup matches the intended experiment. We present the Labware Setup Checker, a passive vision system that verifies deck preparation on an Opentrons Flex without modifying its hardware or firmware. A consumer webcam mounted outside the robot supplies images to a browser application. The checker segments the deck into slots, corrects slot assignments using the known deck geometry, classifies the labware, and compares the observed setup with a reference protocol library to report missing, misplaced, or unexpected labware. It operates independently of the robot controller and does not interrupt or gate a run. During SwabSeq Respiratory Viral Panel (RVP) pre-PCR deck preparation, the checker tracked a slot from empty to a base and then to the completed plate-and-base assembly, identifying the completed assembly in all 29 frames before robot motion. Across these frames, 95.1\% of the 348 labware classifications for slots requiring exact matches agreed with the protocol reference. In 480 frames with partially occluded reservoirs, all 6,720 slot assignments were correct. Across three camera positions and two lighting conditions, slot assignment was 100\% correct, and 97.73\% of labware classifications matched the protocol reference. Median verification time was 2.10 s on a laptop using only its CPU and 4.24 s from frame capture to receipt of results during AWS deployment in the laboratory. These results show that an external camera and independent software can provide verification of deck setup against protocol requirements within seconds, allowing operators to review the physical setup before starting a run.}

\begin{document}

\maketitle

\section{Introduction}

\begin{figure*}[t]
\centering
\begin{minipage}[t]{0.88\linewidth}
\centering
\begin{minipage}[t]{0.50\linewidth}
\centering
{\small\textbf{A. Checker interface}}\\[0.2em]
\includegraphics[width=\linewidth]{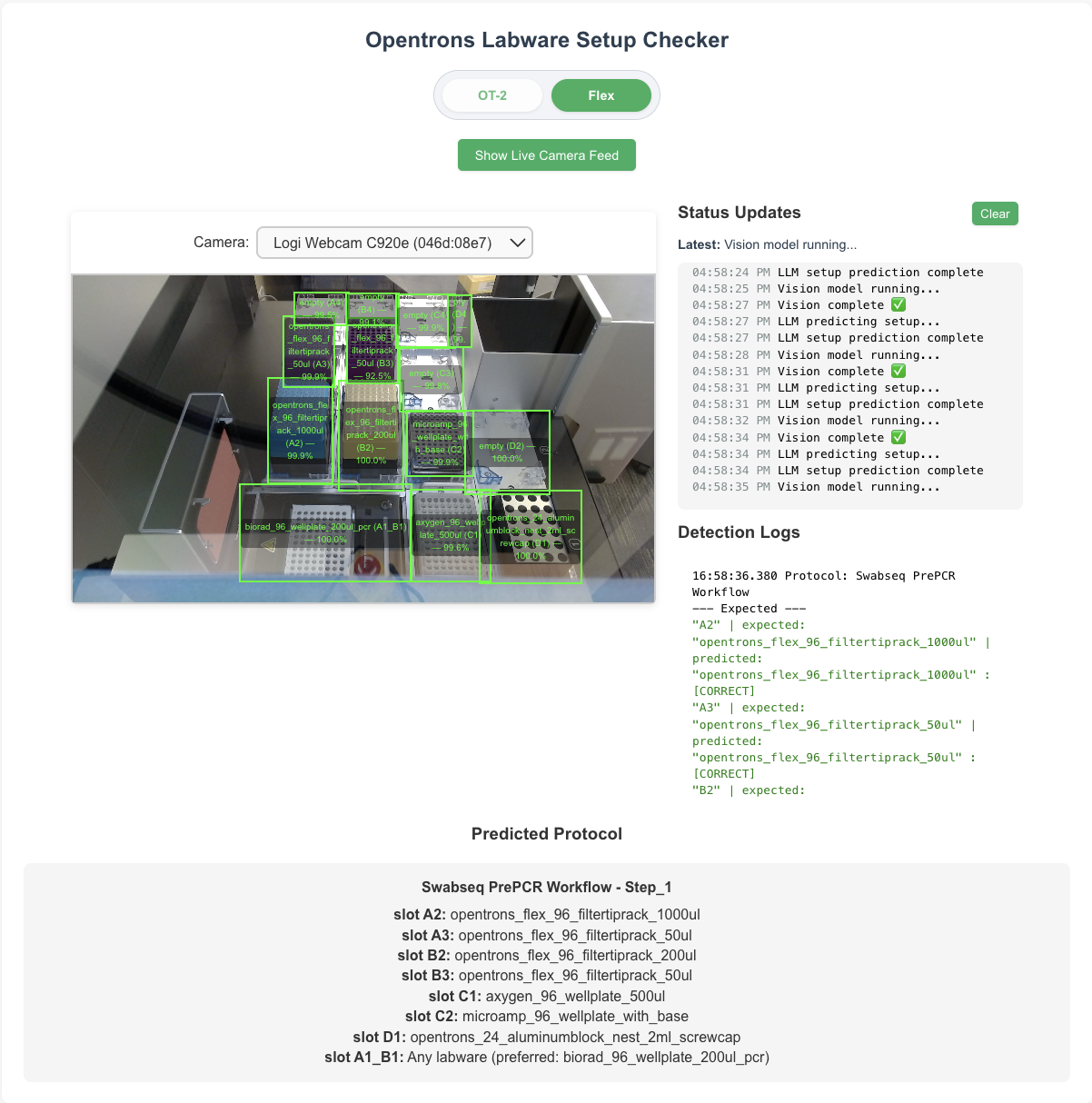}
\end{minipage}\hfill
\begin{minipage}[t]{0.46\linewidth}
\centering
{\small\textbf{B. Incorrect deck setup}}\\[0.2em]
\includegraphics[width=\linewidth]{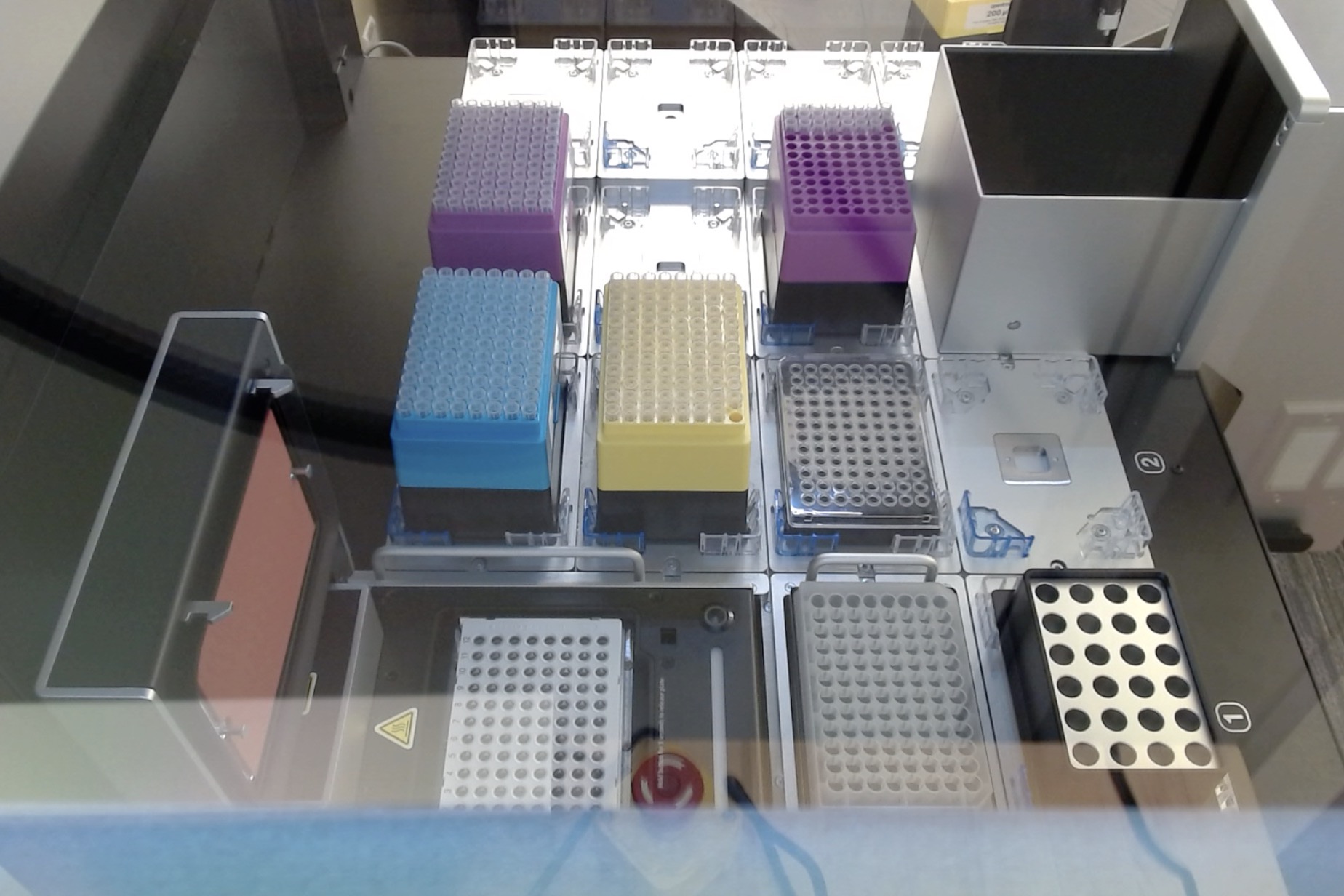}\\[0.5em]
{\small\textbf{C. Returned verification result}}\\[0.2em]
\includegraphics[width=\linewidth]{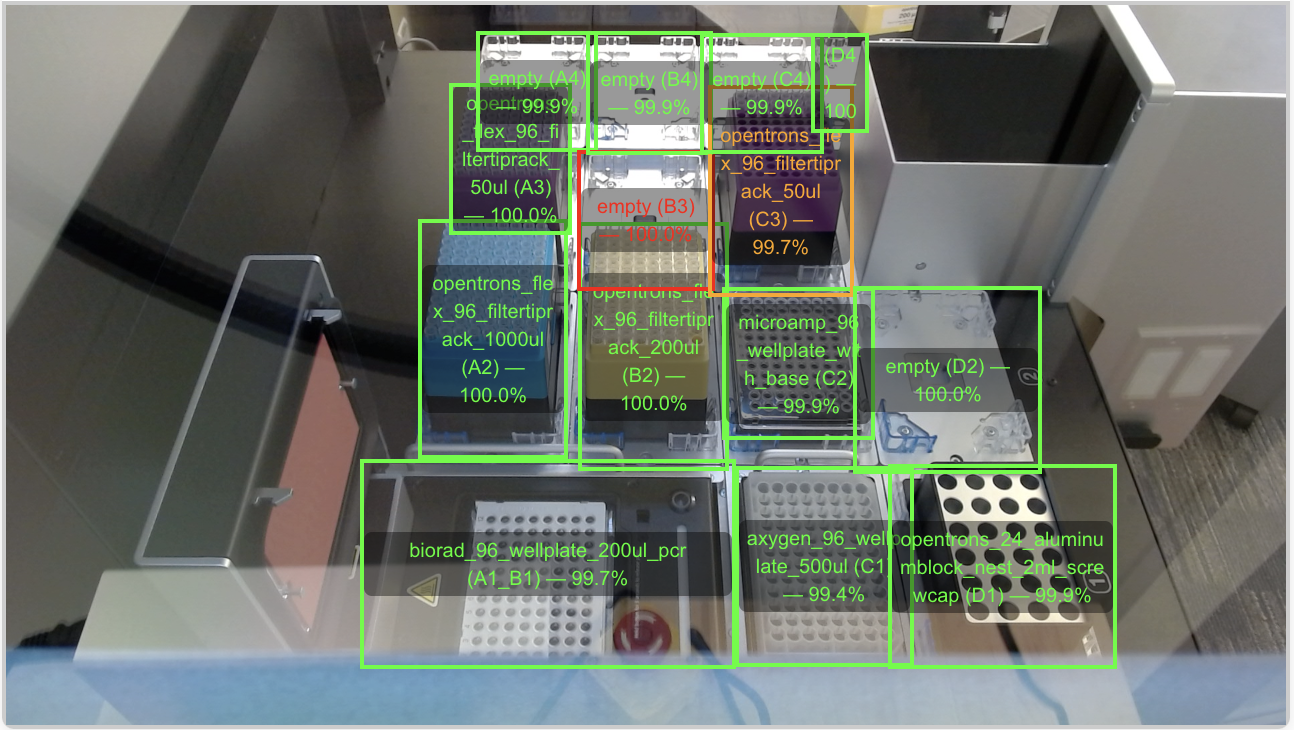}
\end{minipage}
\end{minipage}
\caption{\textbf{Web interface and protocol-aware deck verification.} \textbf{A.} The checker displays the live camera feed with predicted slot and labware labels, status updates, detection logs, and the predicted protocol. \textbf{B.} An incorrect pre-PCR deck setup in which B3 is empty and the expected 50~\textmu L tip rack is placed at C3. \textbf{C.} The returned verification result. Green boxes indicate that the observed labware matches the protocol or that an expected-empty slot is empty. The red box at B3 indicates that the expected tip rack is missing, and the orange box at C3 indicates that the tip rack was detected in the wrong slot.}
\label{fig:webapp}
\end{figure*}

Laboratories are increasingly automating routine workflows. In a 2025 survey, 49\% of respondents reported using robotics or automation in their company's laboratories \cite{Pistoia2025}. Automated liquid handlers are a key tool in laboratory automation, supporting molecular biology workflows such as sequencing library preparation, PCR setup, and high-throughput processing \cite{Schuster2024}. They enable workflows to be run reproducibly with minimal manual labor. These systems are offered by manufacturers such as Opentrons, Tecan, and Hamilton. Quality-control studies of laboratory automation show that liquid handlers require routine checks to stay accurate and reliable, since components can fail over time \cite{Jones2003,Kong2012}. Setup errors, however, can occur even earlier: a plate placed in the wrong slot, the wrong tip rack loaded, a reservoir missing, or labware shifted or occluded on the deck. These errors can cause collisions, damage the robot, or lead to incorrect liquid transfers that compromise the experiment. This physical starting state requires manual assembly: labware, reagents, plates, and tip racks must be placed on the deck in the correct locations before the protocol begins. Correct deck setup relies on user confirmation, and there is no system in place to confirm that the setup matches the robot's expectations, making the initial manual deck setup step prone to errors. Therefore, a monitoring layer is needed to answer three questions: what experiment the operator is running, whether the deck is prepared correctly for it, and whether the robot is executing the protocol correctly.

To implement a monitoring layer, a visual system is required. Currently, camera-based deck checking is available on the Opentrons Flex, which carries a built-in 2 MP camera that records images and video \cite{OpentronsFlexSpecs}. The camera sends images to an operator or to storage, but does not report whether the observed labware is correct for a given protocol. Tecan Fluent workstations run DeckCheck, which photographs the worktable after loading and compares the live layout against a stored reference \cite{TecanFluentManual}. However, DeckCheck only compares pixels and not labware identity, leading to the manual listing differences it cannot flag: different tip types, missing tubes on partially loaded runners, plates rotated by 180 degrees, and well shape \cite{TecanFluentManual}. Research systems have used computer vision to monitor laboratory processes and detect anomalies during automated liquid transfer \cite{ElKhawaldeh2024,Dabouei2025}, and have supported colorimetric titration \cite{Li2025}, but these remain isolated capabilities rather than a general check against a protocol's requirements.

\begin{figure*}[t]
\centering
\includegraphics[width=\linewidth]{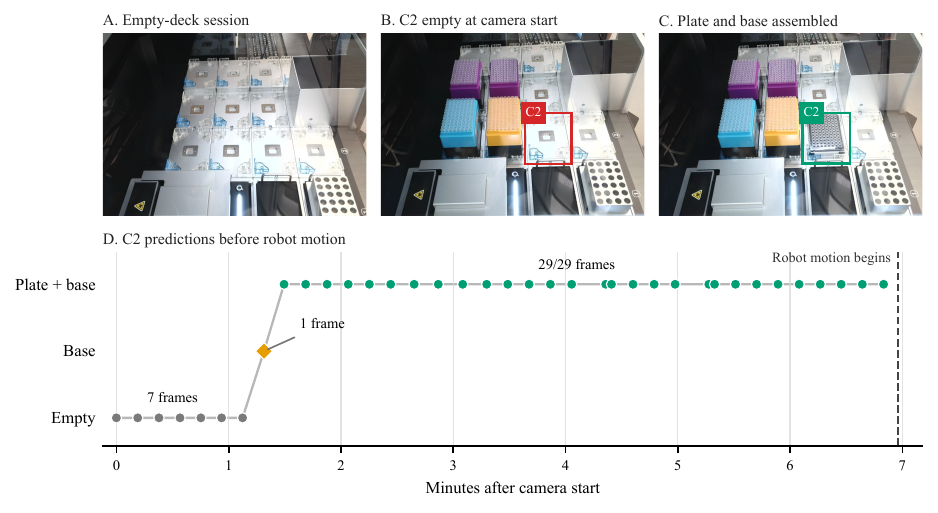}
\caption{\textbf{Recognition of an empty deck and tracking of C2 during SwabSeq Respiratory Viral Panel pre-PCR preparation.} \textbf{A.} Representative frame from the empty-deck session, in which only the permanent C1 temperature module and D1 aluminum block were present. \textbf{B.} C2 was empty when recording began. The red outline indicates that the expected assembly was not yet present. \textbf{C.} The completed MicroAmp plate-and-base assembly at C2. The green outline indicates that the observed assembly matched the protocol. \textbf{D.} C2 predictions across all 37 frames collected before robot motion. Each marker represents one analyzed frame. The checker classified C2 as empty for seven frames, as the MicroAmp base in the next frame, and as the complete MicroAmp plate-and-base assembly in all subsequent 29 frames.}
\label{fig:deckpreparation}
\end{figure*}

Existing options do not provide a modular and accessible verification layer that a laboratory can add to the liquid handler it already owns. DeckCheck is integrated into a high-end workstation and is unavailable for lower-cost platforms. Accessible self-driving laboratories close the loop between planning, execution, measurement, and decision-making \cite{Hase2019,Abolhasani2023,Burger2020}, but also require tightly integrated instruments, data infrastructure, and expertise that academic life-science laboratories often lack \cite{Holland2020}. As a result, a laboratory running an affordable liquid handler has no straightforward way to check its deck against the intended protocol.

Such a system must meet four constraints: it must be inexpensive, leave robot hardware and firmware untouched, run on an available computer, and add no work to the operator's routine. Computer vision makes this checking system practical: a camera is an inexpensive non-contact sensor, and deep learning now localizes, segments, and classifies objects reliably in visually cluttered scenes \cite{LeCun2015,Voulodimos2018}. Applied to a deck image, these methods can find each slot, separate labware from its background, identify the labware type, and map it to the robot's canonical deck coordinates, turning a photograph into a structured deck state that can be compared with a protocol.

We present the Labware Setup Checker, a passive vision system for the Opentrons Flex that utilizes a low-cost camera and lightweight models to confirm an observed deck setup against an intended protocol within seconds. A consumer webcam outside the robot observes the deck. The checker segments the image into slots, classifies masked and cropped labware images, corrects the assignments against canonical slot geometry, and compares the result with a reference protocol library, marking each slot as correct, wrong, misplaced, or unexpected. The checker operates passively and noninvasively, requiring only a simple session setup and checking the deck automatically without interrupting robot operation or adding manual steps to the established procedure. This work contributes an external verification architecture for the Opentrons Flex that requires no hardware or firmware modification, protocol-aware deck perception at seconds-scale latency in a deployed setting, and an evaluation system compatible with routine laboratory workflows.

\section{Results}

\subsection{The checker reports protocol-aware deck states without modifying the robot}

The checker is implemented as a web application (Figure~\ref{fig:webapp}). The operator selects the robot type and camera and then starts the live feed. The checker periodically samples the camera feed and overlays the image with predicted slot labels, labware labels, and confidence scores. It also displays the expected protocol layout and updates the detection log for each sampled frame. Because the checker runs independently of the robot controller, it requires no hardware or firmware modification and does not interrupt robot operation.

\begin{figure*}[t]
\centering
\begin{minipage}{0.75\textwidth}
\centering
\begin{minipage}[t]{0.48\linewidth}
\centering
{\small\textbf{A. Alternate angle 1}}\\[0.2em]
\includegraphics[width=\linewidth]{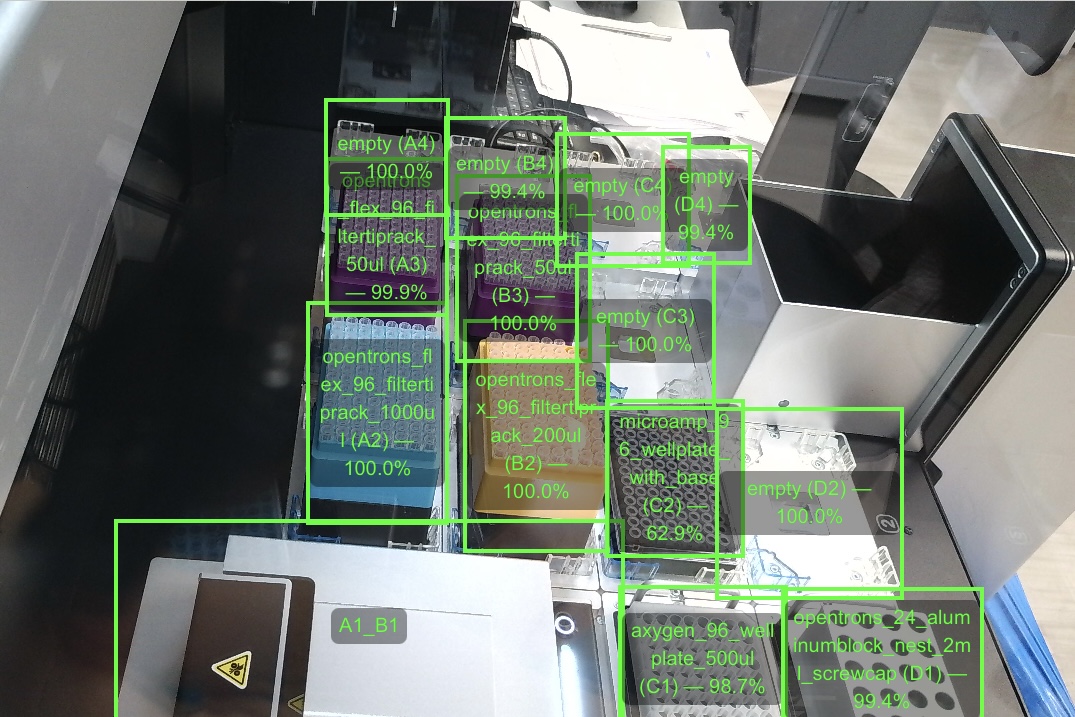}
\end{minipage}\hfill
\begin{minipage}[t]{0.48\linewidth}
\centering
{\small\textbf{B. Alternate angle 2}}\\[0.2em]
\includegraphics[width=\linewidth]{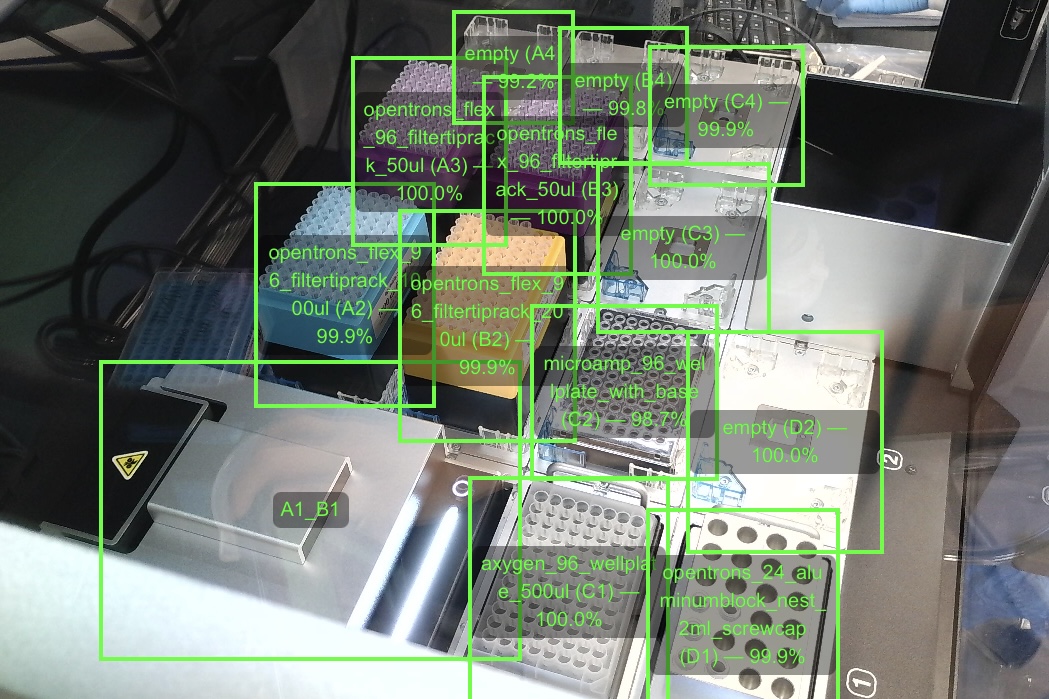}
\end{minipage}
\caption{\textbf{The checker under alternate Flex camera views.} \textbf{A--B.} Two alternate camera viewpoints during live use of the checker.}
\label{fig:edgestress}
\end{minipage}
\end{figure*}

The checker processes each sampled frame by segmenting the observable deck regions, classifying the labware within each region, correcting the assignments against the canonical deck geometry, and comparing the observed slot--labware pairs with a protocol reference. The Opentrons Flex deck was represented by 14 observable regions, including the combined A1/B1 thermocycler region and all visible slots A2 through D4. D3 was excluded because the trash chute occupied that location in the evaluated protocols.

The checker converts the frame into a structured deck state containing the visible slots, predicted labware, classification confidence, and whether each slot matches the expected protocol. Each slot is assigned one of four status labels: \texttt{correct} when the observed state satisfies the protocol reference, \texttt{wrong} when expected labware is missing, \texttt{wrong\_slot} when labware is detected in another protocol slot, and \texttt{not\_in\_setup} when the observed labware is not used by the protocol. The operator can therefore compare the observed deck, expected protocol layout, and per-slot results before starting the run.

\subsection{The checker recognizes an empty deck and tracks deck preparation}

We evaluated the checker on an empty Opentrons Flex deck and during preparation of a SwabSeq Respiratory Viral Panel (RVP) pre-PCR deck, a routine workflow in the laboratory.

To evaluate whether the checker could distinguish empty slots from missed detections, we analyzed 113 frames of an empty Opentrons Flex deck. Only the permanent C1 temperature module and D1 aluminum block remained on the deck. Across all 113 frames, the checker consistently recognized the remaining deck slots as empty (Figure~\ref{fig:deckpreparation}A).

During RVP pre-PCR deck preparation, the checker reported C2 as empty in the first seven frames. After the operator placed the MicroAmp base at C2, the checker identified the base in the next frame. The MicroAmp plate was then placed on the base, and the checker identified the complete MicroAmp plate-and-base assembly correctly in all 29 frames collected before robot motion. The checker therefore tracked the C2 setup from an empty slot to the base and then to the completed assembly (Figure~\ref{fig:deckpreparation}B--D).

For the 29 completed-deck frames, the checker returned results for all 406 slot observations. Of these, 348 observations came from the 12 regions requiring exact labware matches, and the remaining 58 observations matched their expected deck states. Among the 348 exact-match observations, the predicted class matched the known deck setup and protocol reference for 331 (95.1\%); the remaining 17 were misclassified, including 15 at D1 and two at D4. All four tip racks were correctly classified in all 116 observations, the C2 MicroAmp plate-and-base assembly was correctly classified in all 29 observations, and the five expected-empty slots were correctly classified in all 145 observations. Combining the 331 correct exact-match observations with the other 58 correct observations, the checker assigned a \texttt{correct} verification status to 389 of all 406 observations (95.8\%).

\begin{figure*}[t]
\centering
\begin{minipage}{0.8\textwidth}
\centering
\includegraphics[width=\linewidth]{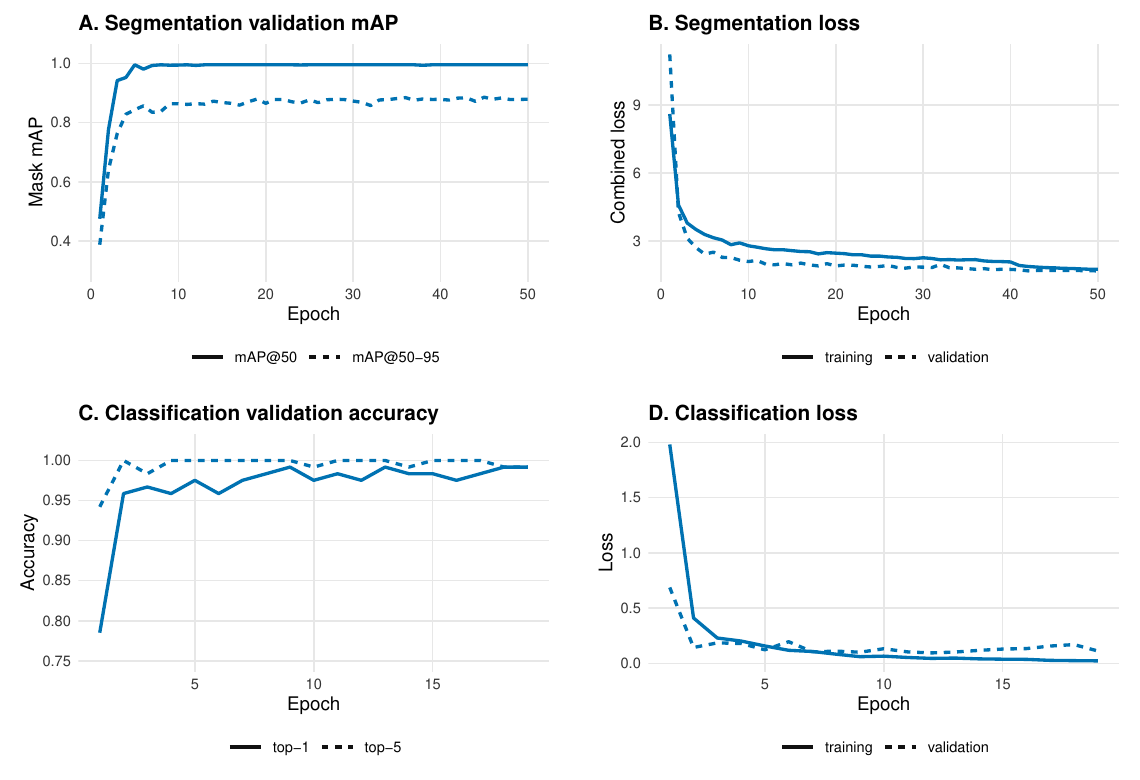}
\caption{\textbf{Training dynamics of the perception models.} \textbf{A.} Segmentation validation performance measured by mask mAP@50 and mAP@50--95 across 50 epochs. \textbf{B.} Segmentation training and validation loss, calculated as the sum of bounding-box, segmentation, classification, and distribution focal losses. \textbf{C.} Classification validation performance measured by top-1 and top-5 accuracy across 19 recorded epochs. \textbf{D.} Classification training and validation loss. Both models converged within the logged training runs, reaching high validation performance while validation loss remained low.}
\label{fig:modelcurves}
\end{minipage}
\end{figure*}

The 17 misclassifications arose from visual similarity between two aluminum blocks at D1 and limited slot visibility at D4. At D1, the expected 24-well screw-cap aluminum block was correctly identified in 14 of 29 frames and classified as a Bio-Rad well-plate aluminum block in the other 15 frames. These two aluminum blocks have similar shapes and surface features, indicating that visually similar labware requires better class-specific training. At D4, the expected empty slot was correctly identified in 27 of 29 frames. In the other two frames, the slot was classified as an Abgene deep-well plate with low confidence. D4 was at the edge of the camera's field of view, and only a narrow portion of the slot was visible.

\subsection{The checker remains robust to occlusion, camera position, and lighting conditions}

To test partial occlusion, we placed four reservoirs with similar appearances across A4, B4, C4, and D4 in all 24 possible arrangements. Tip racks at A3, B3, and C3 partially blocked the reservoirs behind them, while the waste chute obstructed part of D4. We collected 20 consecutive frames for each arrangement, producing 480 frames.

Across the 480 occlusion test frames, deck-region assignment was correct in 6720 of 6720 observations (100\%), covering all 14 regions in every frame. This included correct localization of the four partially blocked reservoir positions in 1920 of 1920 observations (100\%), with 120 observations for each reservoir--position combination. The three foreground tip racks were also classified correctly in 1440 of 1440 observations (100\%). Thus, the oblique view did not prevent the checker from preserving the Flex deck geometry under these intentionally difficult layouts.

We next held the completed RVP pre-PCR deck and protocol reference constant while testing three camera positions under two lighting conditions. The controlled set contained 154 stable frames, with 24 to 28 frames for each camera-position and lighting combination. Figure~\ref{fig:edgestress} shows two alternate camera viewpoints during live use of the checker.

Across the 154 stable camera position and lighting frames, the checker detected and assigned all 14 deck regions in every frame, giving 2156 of 2156 correct deck-region observations (100\%). Across the complete deck, 2107 of 2156 slot classifications matched the protocol reference (97.73\%). The percentages of correct slot classifications at the three camera positions were 96.43\%, 96.71\%, and 99.87\%. The corresponding percentages under the two lighting conditions were 97.68\% and 97.77\%, and the six individual position--lighting combinations ranged from 92.86\% to 100\%.

The 49 incorrect slot classifications followed two main patterns tied to specific camera and lighting combinations. At position 1 under lighting condition 1, the D1 aluminum block was classified as a 15-tube rack in all 25 frames; under lighting condition 2 at the same position, D1 was correct in all 25 frames. At position 2 under lighting condition 2, the B3 50~\textmu L tip rack was classified as an Abgene deep-well plate in 22 of 26 frames; under lighting condition 1, B3 was correct in all 24 frames. One additional D1 error occurred at position 2 under lighting condition 2, and one D4 error occurred at position 3 under lighting condition 2. Because the same labware was classified correctly from the other views, these errors are consistent with reflections changing its visible texture at specific camera and lighting combinations. Overall, the checker achieved 100\% deck-region localization and 97.73\% correct slot classifications across the six combinations.

\subsection{The perception models achieve high validation accuracy}

On the held-out validation sets, the YOLO11m segmentation model reached 0.993 box precision, 1.000 box recall, and 0.995 box mAP@50. Mask mAP@50 was also 0.995, with a mask mAP@50--95 score of 0.884. The YOLO26m classifier reached 0.992 top-1 accuracy and 1.000 top-5 accuracy across 17 classes.

The training logs showed rapid convergence (Figure~\ref{fig:modelcurves}). On an NVIDIA T4 GPU in Google Colab, the segmentation model completed 50 epochs in approximately 19 minutes. Segmentation mask mAP@50 rose sharply during the first several epochs and remained stable through epoch 50. The classifier training record used for Figure~\ref{fig:modelcurves} contains 19 epochs. Its top-1 accuracy remained above 0.95 after epoch 2 and ended at 0.992. A separate classifier training run completed in approximately 15 minutes.

\subsection{The checker completes deck verification locally and in deployment within seconds}

The checker completed deck verification within seconds both locally and during deployment of the TECAN Library Cleanup workflow. Across 10 post-warmup runs on the CPU-only development MacBook Pro, the complete Flex deck-verification path processed a representative \(2304\times1536\) pixel image in a median of 2.10 s. Preprocessing and PNG serialization took 1.24 s, segmentation took 0.624 s, and classification of 14 masked cropped images took 0.188 s. Slot correction and protocol matching contributed little to the total latency.

The same trained models were used for RVP pre-PCR deck preparation and the TECAN Library Cleanup workflow without additional training or protocol-specific validation. Supporting the additional workflow required only adding its expected deck setup to the backend protocol library. The deployed latency evaluation included 280 frames processed by the AWS-hosted checker during a TECAN Library Cleanup workflow. The checker processed each frame on the server in a median of 3.2 s. Median stage times were 1.64 s for preprocessing, 1.15 s for segmentation, 0.36 s for classification, 26 ms for slot correction, and 1.3 ms for protocol matching. Median end-to-end latency from frame capture to receipt of the verification result was 4.24 s (IQR, 3.52--4.48 s). The median estimated network and browser overhead was 1.10 s. In both the local and deployed settings, image preprocessing contributed more latency than model inference or protocol matching. Although the checker was not optimized for speed, the measured latency was short relative to the several-minute deck-preparation period.

\section{Discussion}

We developed the checker to convert images of an Opentrons Flex deck into structured verification results for each slot, allowing for independent and external validation of proper deck preparation. The checker identifies the observed labware, compares the deck with a reference protocol, and reports whether each slot is correct, empty, misplaced, or unexpected. Its purpose is limited to verifying deck preparation, and it does not plan experiments, control the robot, or monitor liquid handling operations.

The checker was designed to simplify deployment of routine molecular biology workflows. It uses a consumer webcam and browser interface, operates independently of the Flex controller, and requires no modification to the robot hardware or firmware. Complete deck verification took a median of 2.10 seconds on a laptop using only its CPU. The AWS-hosted checker also returned results within seconds during laboratory operation. Tests across three camera positions and two lighting conditions showed complete deck region localization and high agreement with the protocol reference under the evaluated conditions.

Operating independently of the controller creates a deliberate tradeoff. The checker cannot prevent an operator from starting a run after it reports a mismatch. Instead, it provides an additional verification step that allows the operator to inspect and correct the deck before proceeding with a protocol. This separation simplifies deployment and preserves operator control, but the operator must notice and respond to the displayed result. Commercial deck verification systems use a similar approach by allowing operators to ignore a warning and continue \cite{TecanFluentManual}.

Adoption in another laboratory requires configuration. A new protocol must be represented as an expected arrangement of labware across the deck. The camera must be positioned so that every required deck region is sufficiently visible. Labware already represented in the classifier can be used in new protocol layouts. Unseen or visually different labware requires image collection, annotation, model retraining, and validation. The current checker is therefore configurable, but it is not ready for immediate use with every protocol and labware type.

Several limitations define the scope of the results. The laboratory evaluation was conducted on Flex robots in one laboratory and covered a limited set of workflows. Agreement with the protocol reference is not the same as independently annotated accuracy for detecting setup errors. A reported mismatch may represent either a model error or a genuinely incorrect deck. Classification can also fail when labware types look similar or when their appearance is not represented in the training data. Reliable verification also requires adequate camera coverage. Finally, the checker identifies the deck state before a run but does not measure liquid volumes, detect individual missing tips or tubes, or monitor pipetting and other failures during a run.

Future work should generate protocol references directly from Opentrons protocol files, simplify the addition of new labware, and evaluate the checker across more laboratories, camera installations, and workflows. Recording genuine setup mistakes would help determine how often the checker detects errors that require action before a run. Future versions could also use the Flex camera or communicate with the robot controller. The checker could be extended to monitor labware presence and position during experiments and record these observations in an electronic lab notebook (ELN). Linking time-stamped observations to the protocol could help scientists document experimental progress and review deviations from the expected setup. Overall, these results show that protocol aware deck verification can be added to an existing liquid handler without requiring direct robot control.

\section{Methods}

We built the checker using an external webcam, YOLO11m segmentation and YOLO26m classification models, an inference and protocol matching pipeline, and a browser application deployed on AWS. The following sections describe the hardware and data collection, model training, inference, protocol matching, AWS deployment, and evaluation.

\subsection{Hardware}

\begin{figure}[t]
\centering
\includegraphics[width=\linewidth]{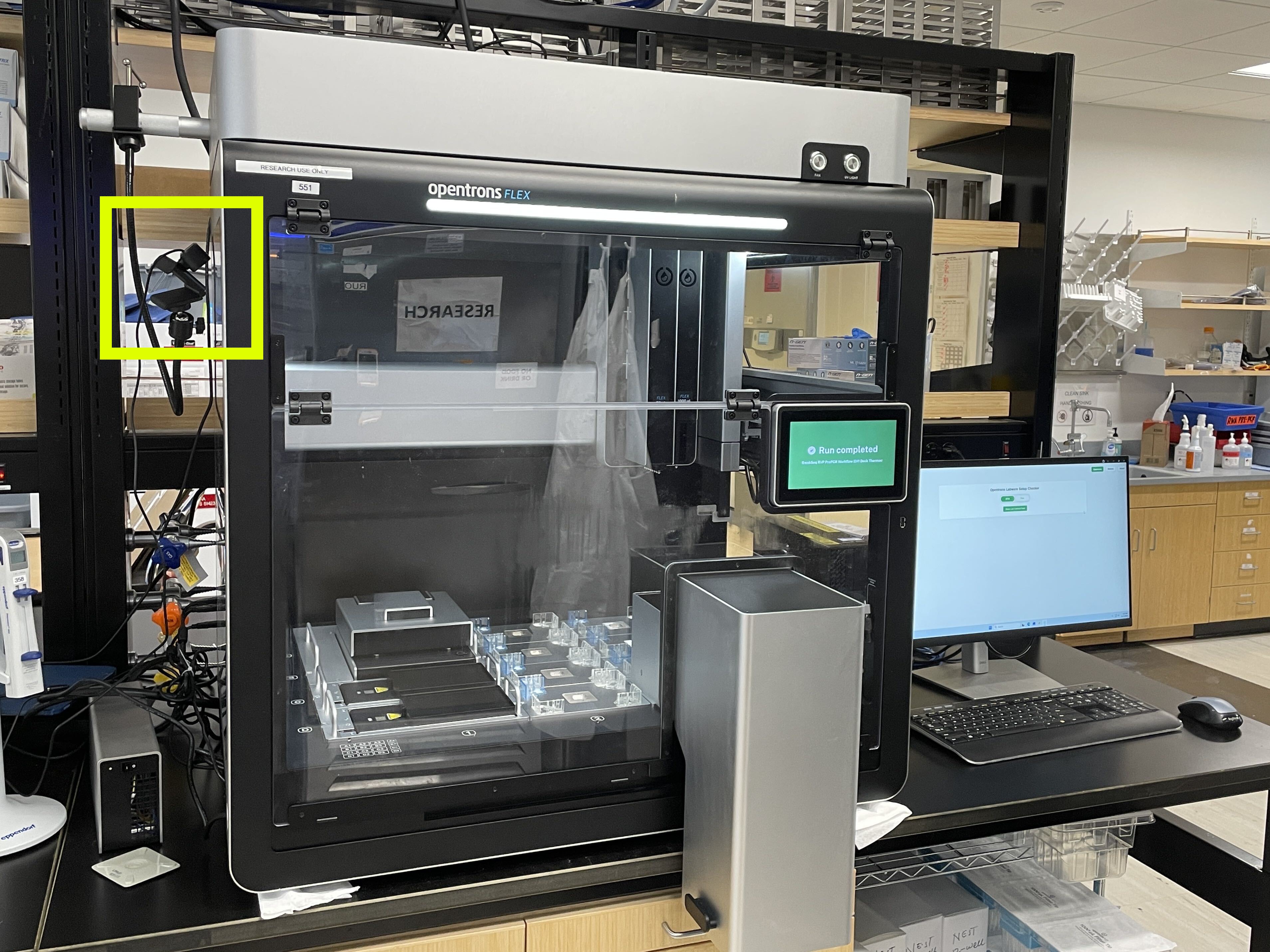}
\caption{\textbf{Laboratory deployment of the checker.} A consumer webcam (yellow box) mounted beside the Opentrons Flex observes the deck, while the browser interface runs on an adjacent computer. The camera and computer operate outside the robot controller and require no modification to the Flex hardware or firmware.}
\label{fig:labdeployment}
\end{figure}

We deployed the checker on an Opentrons Flex liquid handler. The primary camera was a Logitech C920e 1080p webcam, a low-cost consumer camera that retailed for roughly \$60 from Amazon at the time of writing, mounted on an adjustable stand outside the robot body (Figure~\ref{fig:labdeployment}). The primary stand was an InnoGear webcam desk mount stand, retailing for around \$20 from Amazon at the time of writing. The stand was attached to an Opentrons Flex lift handle screwed into the designated location in the Opentrons Flex HEPA/UV Module. The lift handle was included with the robot for easy relocation. Camera placement was chosen to maximize visibility of the deck while avoiding interference with gantry motion, pipette access, and normal operator setup. The Flex deployment used an upper-left external viewpoint to reduce occlusion from the waste chute and side-mounted accessories.

\subsection{Dataset collection and annotation}

\begin{figure*}[t]
\centering
\includegraphics[width=\linewidth]{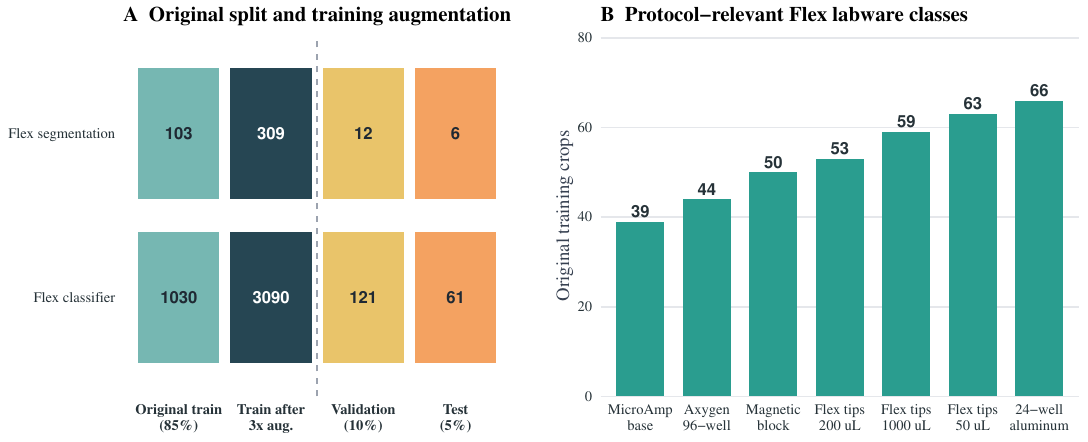}
\caption{\textbf{Dataset composition for model development.} Panel A shows the 85/10/5 train/validation/test split and the threefold augmentation applied to the training split only; teal columns are training data and gold/orange columns are held-out validation and test data. Counts are full-deck images for segmentation datasets and cropped labware images for classification datasets. Panel B shows original, pre-augmentation Flex classifier cropped training images for labware classes used by the prePCR and postPCR protocols. Empty slots and rare non-protocol classes are not included in Panel B.}
\label{fig:datasetcoverage}
\end{figure*}

Training photographs were captured from real Flex decks using Photo Booth on a MacBook under laboratory lighting conditions. We collected 121 full-deck images with randomized labware arrangements. Labware was placed in different deck positions so that the models would not learn a fixed association between labware type and slot. The full-deck images had average dimensions of approximately \(1760\times1178\) pixels. We manually annotated the visible labware and slot regions on the Roboflow platform. Empty slots were explicitly labeled so that the models could distinguish an empty position from a missed detection. Individual labware and empty-slot regions were cropped to create a classification dataset containing 1,212 images across 17 classes (Figure~\ref{fig:datasetcoverage}B), including empty slots, tip racks, reservoirs, well plates, tube racks, aluminum blocks, and magnetic blocks. The cropped images had average dimensions of approximately \(458\times409\) pixels.

\subsection{Model training}

We divided each dataset into training, validation, and test sets using an 85/10/5 train/validation/test split. The segmentation dataset was divided into 103 training images, 12 validation images, and 6 test images. The classification dataset was divided into 1,030 training images, 121 validation images, and 61 test images. We augmented only the training sets on the Roboflow platform. Augmentation included horizontal and vertical flips, \(90^\circ\) clockwise and counterclockwise rotations, rotations between \(-15^\circ\) and \(+15^\circ\), brightness adjustments between \(-15\%\) and \(+15\%\), and exposure adjustments between \(-10\%\) and \(+10\%\). These augmentations expanded the segmentation training set from 103 to 309 images and the classification training set from 1,030 to 3,090 images (Figure~\ref{fig:datasetcoverage}A).

\begin{figure*}[t]
\centering
\includegraphics[width=0.88\linewidth]{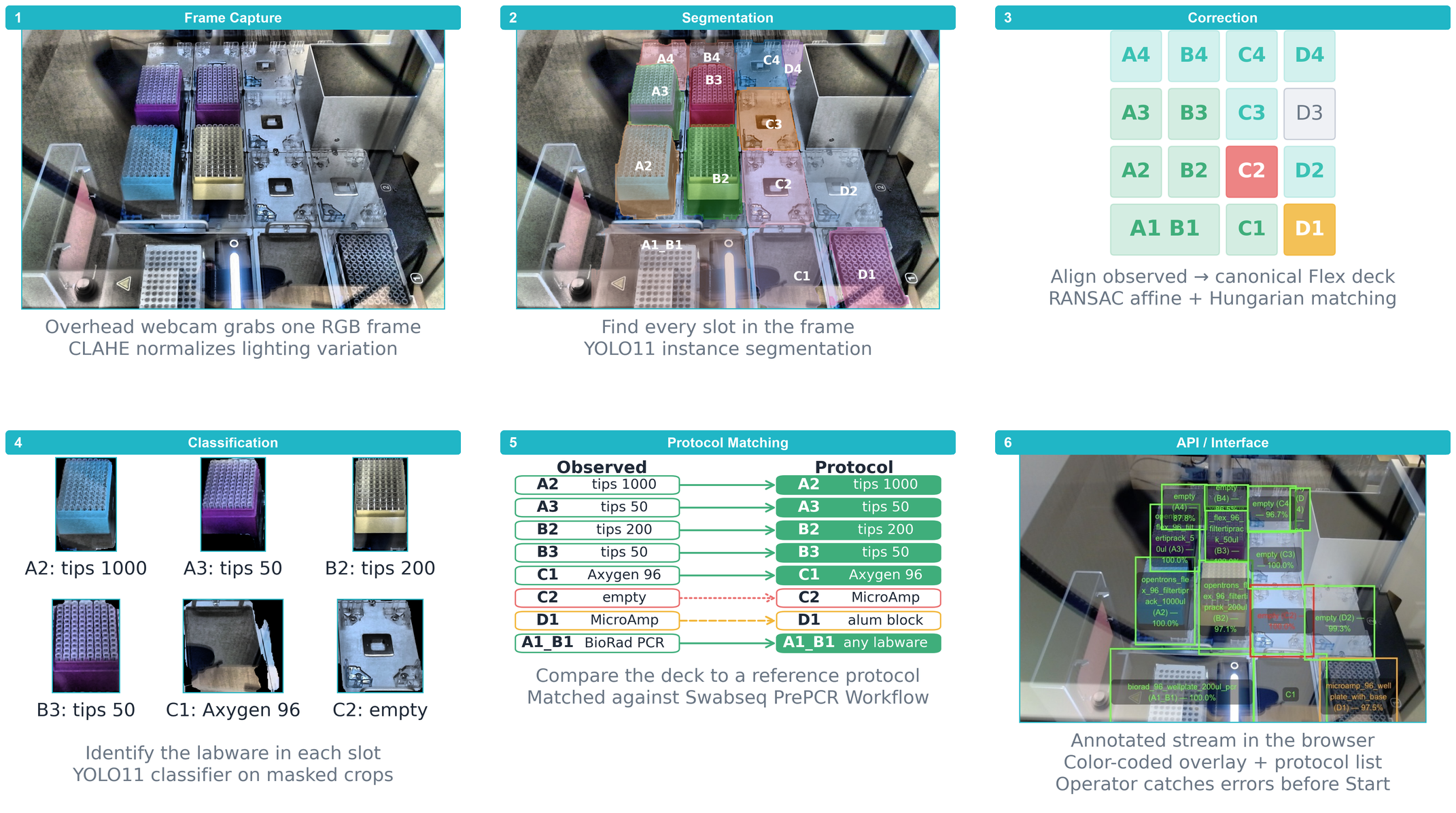}
\caption{\textbf{Computer vision and protocol-matching pipeline.} A camera frame is normalized, segmented into deck slots, geometrically corrected to the canonical Flex deck, classified with masked and cropped labware images, compared against the reference protocol, and returned to the browser as an annotated interface.}
\label{fig:methodpipeline}
\end{figure*}

We trained a YOLO11m segmentation model and a YOLO26m classification model using the Ultralytics training framework \cite{Redmon2016,UltralyticsYOLO11}. The segmentation model used \(640\times640\) pixel inputs. The full-deck images retained their collected resolution, and the Ultralytics training pipeline converted them to the configured \(640\times640\) input size during data loading. The classification model used \(224\times224\) pixel inputs, and Roboflow resized the cropped classification images to \(224\times224\) pixels using stretch resizing. The segmentation model detected 14 Flex slot classes, including the combined A1/B1 region for the thermocycler module and the visible deck slots A2 through D4. D3 was excluded because the trash chute occupies that location and was the only equipment placed there in the SwabSeq protocols evaluated in this study. The segmentation model was configured for 50 epochs with a batch size of 16 and early stopping patience of 10. The classification model was configured for 30 epochs with a batch size of 32 and early stopping patience of 10. Both models were trained on an NVIDIA T4 GPU in Google Colab. The trained model weights were exported in OpenVINO intermediate representation format for CPU inference \cite{IntelOpenVINO}.

\subsection{Inference}

Figure~\ref{fig:methodpipeline} summarizes the inference and protocol-matching pipeline applied to each camera frame.

\paragraph{Image preprocessing.} Each camera frame was converted to RGB and represented as unsigned 8-bit pixel values. We applied contrast-limited adaptive histogram equalization with a clip limit of 0.03 to reduce sensitivity to lighting variation \cite{Zuiderveld1994}. The frame was then letterbox-resized to the segmentation model input size. This transformation preserved the aspect ratio and recorded the scale and padding offsets used to map model outputs back to the original image.

\paragraph{Deck-region segmentation.} The segmentation model returned a bounding box, class score, and mask coefficients for each detected deck region. We removed detections below the confidence threshold and applied class-wise non-maximum suppression to remove duplicates. Binary instance masks were reconstructed from the prototype masks and detection-specific coefficients, mapped to the original image dimensions, and thresholded. Applying each mask to its corresponding cropped image reduced the contribution of neighboring objects and background regions to labware classification.

\paragraph{Canonical slot assignment and labware classification.} Perspective distortion and occlusion can cause the segmentation model to assign a detected region to a neighboring Flex slot. We corrected these assignments using the known deck geometry. Detection centers were aligned with canonical slot coordinates by estimating an affine transformation with RANSAC \cite{FischlerBolles1981}. The Hungarian algorithm then assigned detections to slots using a cost that combined geometric distance, consistency with the original segmentation label, and detection confidence \cite{Kuhn1955}. When a reliable affine transformation could not be estimated, the checker used a fallback transformation based on the observed bounding-box spans. The Flex labware classifier analyzed the masked and cropped image assigned to each corrected slot and returned one of 17 labware classes. Model logits were converted to probabilities, and the class with the highest probability was returned with its confidence score. Empty slots were treated as a labware class so that the checker could distinguish an expected empty position from a missed detection.

\subsection{Protocol matching}

\begin{figure*}[t]
\centering
\includegraphics[width=0.95\linewidth]{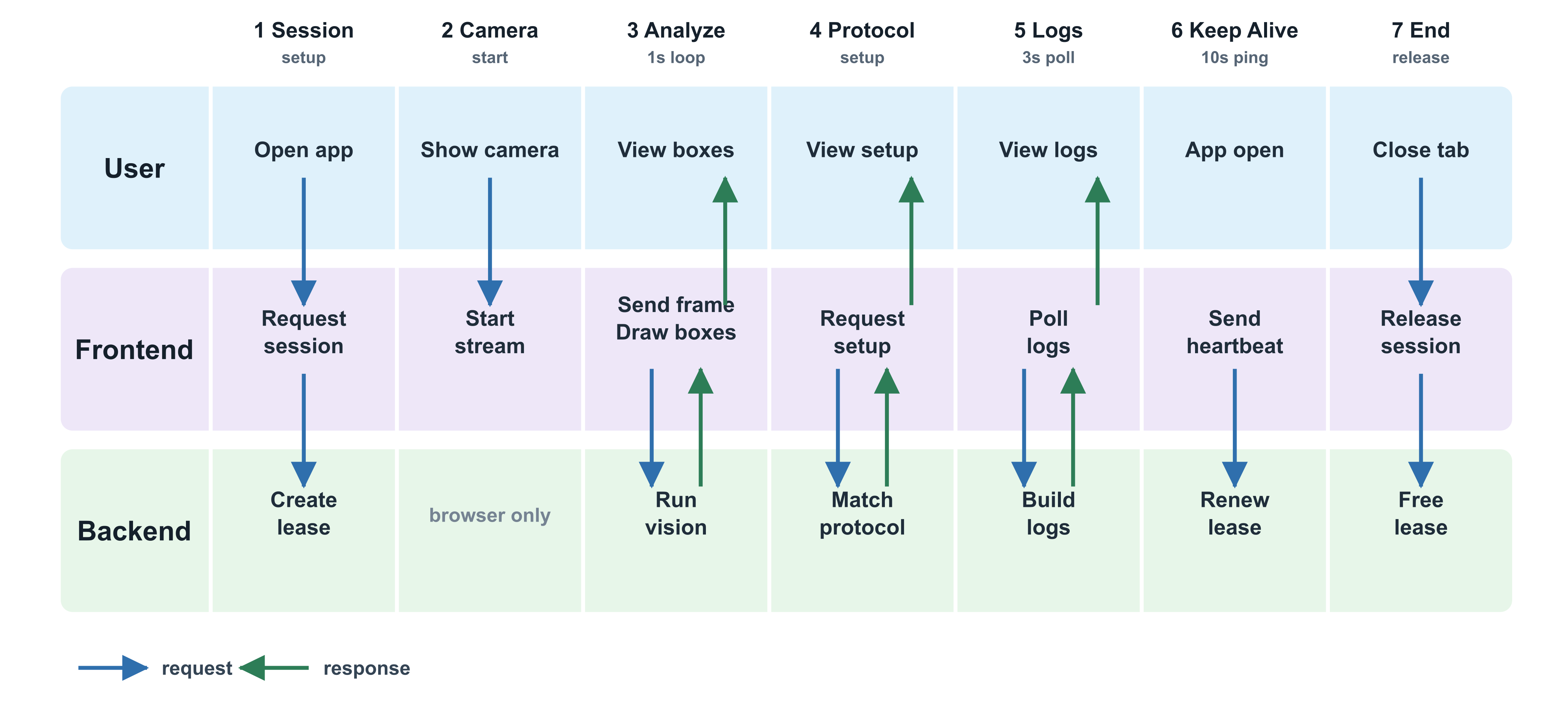}
\caption{\textbf{Checker communication flow.} The browser frontend manages the user session, camera stream, periodic frame analysis, protocol setup requests, log polling, keep-alive messages, and session release. The FastAPI backend creates the session lease, runs the vision and protocol-matching pipeline, builds logs, renews the lease, and frees backend state when the user closes the checker.}
\label{fig:methodarchitecture}
\end{figure*}

The checker compared the observed slot--labware pairs with each layout in the reference protocol library. Exact slot--labware matches received the highest score, labware found in a different protocol slot received partial credit, and permissive slots accepted any observed labware. The checker selected the highest-scoring protocol and returned its expected slot labels. Each observed slot was then assigned a status: \texttt{correct} for an exact or allowed match, \texttt{wrong} when a slot expected to be occupied appeared empty, \texttt{wrong\_slot} when the observed labware belonged elsewhere in the protocol, and \texttt{not\_in\_setup} when the observed labware was not used by the protocol.

\subsection{AWS deployment}

The backend of the checker was a Python FastAPI server containerized for Amazon Web Services (AWS) deployment with Docker, and the frontend was a React single-page application hosted on AWS Amplify and served to the operator's browser. In the laboratory deployment, the server image was pushed to AWS Elastic Container Registry and ran as an AWS-hosted service; session state and recent detections were stored in AWS ElastiCache for Valkey, while captured frames and masked images were written to Amazon S3 for audit, debugging, and further model training. During a session, the camera stream was accessed by the browser through the MediaDevices API. The frontend captured frames from the camera and posted them to the backend, which ran preprocessing, segmentation, slot correction, classification, and protocol matching (Figure~\ref{fig:methodarchitecture}). The backend returned detections, bounding boxes, labels, confidences, and protocol metadata as a JSON file.

\subsection{Evaluation}

Model level performance was evaluated using the held out validation sets produced by the training pipeline. For the Flex segmentation model, we report precision, recall, mAP@50, and mAP@50 to 95 for bounding boxes and masks. For the Flex labware classifier, we report top 1 and top 5 accuracy. Training curves were examined to assess convergence across the completed training runs.

To evaluate deck state recognition during laboratory operation, we recorded an empty Flex deck and the preparation of a SwabSeq Respiratory Viral Panel (RVP) pre-PCR deck \cite{Bloom2021,Kang2022}. The empty deck configuration contained only the permanent C1 temperature module and D1 aluminum block, and 113 frames were analyzed. For the pre PCR preparation test, recording began approximately seven minutes before the robot moved and continued while the operator completed the deck. The four tip racks were already present when recording began, while slot C2 progressed from empty, to a MicroAmp base, and then to the complete MicroAmp plate-and-base assembly. Frames collected after robot motion began were excluded from the pre run analysis because moving robot components could obstruct the deck. We calculated the percentage of correct slot classifications from the status returned by the checker for each slot in each analyzed frame. A slot was counted as correct when the observed labware matched the protocol reference or when the slot was configured as permissive by the reference library. We report this percentage across the complete 14 region Flex deck and across the 12 regions requiring an exact match. For the completed pre-PCR deck, the evaluation included the 29 frames recorded after the complete MicroAmp plate-and-base assembly was placed at C2 and before robot motion began.

To determine whether the checker could reconstruct a complete Flex deck from an oblique view with partial occlusion, we tested deliberately difficult labware arrangements. The upper left camera view creates a difficult case when tall labware in row 3 blocks the staging positions behind it, which include slots A4, B4, C4, and D4. Robustness to partial occlusion was evaluated using four reservoirs with similar appearances: a 15~mL NEST 12 channel reservoir, a 90~mL Axygen single well reservoir, a 195~mL NEST single well reservoir, and a 290~mL Agilent single well reservoir. The reservoirs were placed across A4, B4, C4, and D4 in all 24 possible arrangements, so each reservoir appeared six times in each position. A 200~\textmu L tip rack at A3, a 50~\textmu L tip rack at B3, and a 1000~\textmu L tip rack at C3 partially occluded the reservoirs behind them, while the waste chute obstructed part of D4. We collected 20 consecutive frames for each arrangement, producing 480 frames. Deck region assignment was evaluated across all 14 regions, and reservoir localization was summarized separately for the labware in row 4. Classification of the three foreground tip racks was also evaluated separately.

Robustness to camera placement and lighting was evaluated using the same completed SwabSeq RVP pre-PCR deck at three camera positions under two lighting conditions, producing six combinations of camera position and lighting. Position 1 placed the camera in the middle, position 2 placed it at the left corner of the Flex, and position 3 placed it at the right corner. Each position provided an oblique view of the complete deck. Lighting condition 1 used the brighter, directly illuminated deck view, whereas lighting condition 2 used ambient laboratory illumination without direct deck lighting. The controlled sequence contained 154 stable frames, with 24 to 28 frames per combination. 57 preliminary frames acquired before the controlled sequence and 8 frames captured while the camera was being repositioned were excluded. The deck configuration and reference protocol were held constant. For each stable frame, we measured deck region detection, canonical slot assignment, and the percentage of slot classifications that matched the protocol reference. Results were summarized for each camera position, lighting condition, combination of camera position and lighting, and deck position.

Pipeline latency was measured on the development MacBook Pro using a representative Flex test image at \(2304\times1536\) pixel resolution. Models were initialized once. We then ran the complete pipeline once and excluded that run from the timing analysis. We measured the following 10 complete runs of the same analysis path used by the checker: image decoding, contrast limited adaptive histogram equalization, PNG serialization, segmentation, Flex slot correction, sequential classification of all corrected cropped images, and protocol matching. We report median times for each stage and for the complete pipeline from these runs. Deployment latency was evaluated using 280 frames processed by the AWS hosted checker during a TECAN Library Cleanup workflow. The backend recorded the duration of frame decoding, preprocessing, segmentation, slot correction, classification, protocol matching, metadata construction, state persistence, and the complete server request. Classification time was summed across the sequential classifier calls for each slot. Image and metadata uploads that continued after the verification result was returned were excluded because they did not affect the operator's waiting time. The browser recorded timestamps immediately after capturing each frame and when the corresponding server response arrived. End to end latency was calculated as the difference between these timestamps, and network and browser overhead was estimated by subtracting server processing time from end to end latency. We report median latency for each stage and the median and interquartile range of end to end latency.

\section*{Data and code availability}

The data and code supporting this study will be released publicly, with access details provided in a revised version.

\section*{Funding}

Funding information will be provided in a revised version.

\bibliographystyle{unsrtnat}
\bibliography{sample}

\end{document}